 \documentclass[pmlr,twocolumn,10pt]{jmlr} 

\usepackage{booktabs}
\usepackage{siunitx}
\usepackage{cleveref}
\usepackage{dsfont}
\usepackage{gensymb}
\usepackage{kantlipsum}
\usepackage{bbm}

\usepackage{amsmath, amssymb}
\usepackage{algorithm2e}
\usepackage{array,graphicx}
\usepackage{caption}
\usepackage{longtable}
\DeclareMathOperator*{\argmax}{arg\,max}

\theorembodyfont{\upshape}
\theoremheaderfont{\scshape}
\theorempostheader{:}
\theoremsep{\newline}

\jmlrvolume{225}
\jmlryear{2023}
\jmlrsubmitted{LEAVE UNSET}
\jmlrpublished{LEAVE UNSET}
\jmlrworkshop{Machine Learning for Health (ML4H) 2023} 

\title[Compositional Q-learning]{Compositional Q-learning for electrolyte repletion with imbalanced patient sub-populations}

\author{%
\Name{Aishwarya Mandyam}
\Email{am2@stanford.edu}\\
\addr Stanford University
\AND
\Name{Andrew Jones}\Email{ajones788@gmail.com}\\
\addr Princeton University
\AND
\Name{Jiayu Yao}\Email{jiayu.yao@gladstone.ucsf.edu}\\
\addr Gladstone Institutes
\AND
\Name{Krzysztof Laudanski}\Email{krzysztof.laudanski@uphs.upenn.edu}\\
\addr University of Pennsylvania
\AND
\Name{Barbara E. Engelhardt} \Email{barbarae@stanford.edu}\\
\addr Gladstone Institutes, Stanford University
}

\begin{document}

\maketitle
\begin{abstract}
Reinforcement learning (RL) is an effective framework for solving sequential decision-making tasks. However, applying RL methods in medical care settings is challenging in part due to heterogeneity in treatment response among patients. Some patients can be treated with standard protocols whereas others, such as those with chronic diseases, need personalized treatment planning. Traditional RL methods often fail to account for this heterogeneity, because they assume that all patients respond to the treatment in the same way (i.e., transition dynamics are shared). We introduce Compositional Fitted $Q$-iteration (CFQI), which uses a compositional task structure to represent heterogeneous treatment responses in medical care settings. 
A compositional task consists of several variations of the same task, each progressing in difficulty; solving simpler variants of the task can enable efficient solving of harder variants. 
CFQI uses a compositional $Q$-value function with separate modules for each task variant, allowing it to take advantage of shared knowledge while learning distinct policies for each variant. We validate CFQI's performance using a Cartpole environment and use CFQI to recommend electrolyte repletion for patients with and without renal disease. Our results demonstrate that CFQI is robust even in the presence of class imbalance, enabling effective information usage across patient sub-populations. CFQI exhibits great promise for clinical applications in scenarios characterized by known compositional structures.
\end{abstract}
\begin{keywords}
compositional Q-learning, electronic health records, clinical decision support
\end{keywords}

\section{Introduction}
\label{sec:intro}
In the healthcare domain, patient care settings can be naturally described as sequential decision-making tasks in which an agent, usually representing a medical practitioner, aims to treat each patient effectively.
Recently, reinforcement learning (RL) methods have gained attention for their use in healthcare applications. In these settings, an RL agent is trained to make treatment recommendations that maximize a pre-defined treatment objective.
In healthcare applications, patient information is often documented in electronic health records (EHRs), which contain information ranging from clinical notes to lab measurements. 
By leveraging the wealth of information in EHR datasets, RL has shown success in guiding critical medical decisions, such as glucose control in diabetic patients~\citep{glucose_control}, disease diagnosis~\citep{Kao_Tang_Chang_2018}, and chemotherapy administration~\citep{chemotherapy}. 

\begin{figure}[ht!]
\includegraphics[width=\linewidth]{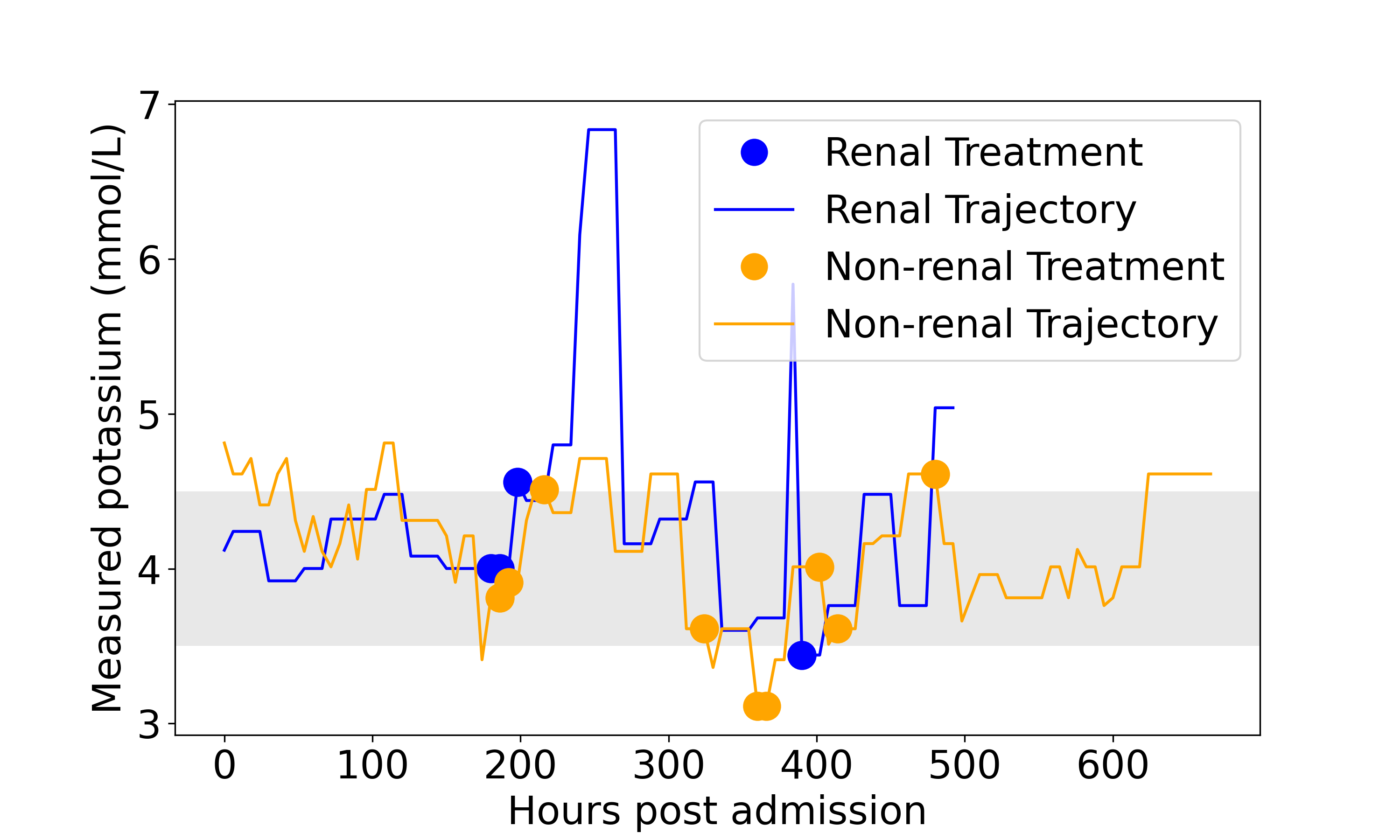}
  \caption{\textbf{Measured blood potassium levels for two example patients: one renal and one non-renal.} Both patients are treated for electrolyte repletion, where the goal is to keep potassium in a reference range (gray shading). The x-axis represents the hours post admission and the y-axis represents the measured potassium level. 
  The blue trajectory represents the potassium value of a renal patient while the orange one represents that of a non-renal patient.
  The blue and orange dots represent electrolyte repletion treatment for renal and non-renal patients respectively. 
 The renal patient has wider deviations from the reference range (more spikes in potassium levels) than the non-renal patient. }
  \label{fig:abstractfigure}
\end{figure}

However, applying RL algorithms to EHR datasets is challenging. Patient populations within EHR are diverse and exhibit heterogeneous treatment responses. 
For instance, patients with chronic diseases often need more careful treatment planning because they respond differently to treatments than patients without chronic diseases.
As an example, consider an electrolyte repletion task in 
the Intensive Care Unit (ICU). Here, the treatment objective is to maintain measured blood potassium levels within a reference range through intravenous (IV) or oral supplementation.
In this scenario, the patients can be divided into two groups: one group with healthy kidney function (non-renal patients) and another group with end-stage renal disease (renal patients), with patients from the two groups requiring different treatment strategies. To see this phenomenon in action, consider the potassium trajectories of example renal and non-renal patients from 
MIMIC-IV~\citep{mimicivdataset}, a large publicly available EHR dataset from the Beth Israel Deaconess Medical Center. The reference range of blood potassium is shared across both groups, and treatments are prescribed when the potassium level has a downward trend. In contrast to the non-renal patient, the renal patient experiences a spike in potassium level after the treatment is administered due to their kidney's compromised filtering capacity (Figure~\ref{fig:abstractfigure}). For renal patients, high potassium levels can lead to severe adverse effects. Thus, clinicians often prescribe small doses of potassium for renal patients and closely monitor the patients' responses with longer intervals between administrations~\citep{esrd_treatment}.

In addition to patient heterogeneity, EHR datasets often exhibit large sample-size imbalance across patient sub-populations. In the ICU setting described earlier, renal patients make up 6\% of the cohort, with non-renal patients comprising the remaining 94\%. Learning a single RL policy in this imbalanced scenario would assume uniform treatment effects across both patient groups and likely ignore the unique treatment responses of renal patients due to their severe underrepresentation in the data.
Moreover, training entirely separate policies for the two patient groups could lead to poorly performing treatment plans for renal patients due to small corresponding sample size.

To tackle the challenge of patient heterogeneity and sample imbalance, we frame policy learning with heterogeneous treatment effects as a compositional RL task.
A compositional RL task is one comprised of several variants of the same task. These variants range in degree of difficulty, and learning how to solve simpler variants enables the agent to solve more complex variants. 
In this setting, we assume that there exists a simpler background variant and one or more foreground variants that are more complex than the background and comparable in complexity to each other.
In the electrolyte repletion
example, the background variant is treating non-renal patients, and the foreground variant is treating renal patients. Administering electrolyte repletion to renal patients is more challenging because they experience spikes in electrolyte levels that are not observed in non-renal patients. 
In a compositional RL task, we assume that we can decompose a foreground variant into the background variant and an additional component that captures 
the difference in treatment 
required to solve the foreground variant.
By leveraging the insights derived from solving the background variant, the foreground variants can be rendered more feasible even with a smaller sample size. 

In this paper, we propose a compositional $Q$-learning based approach that explicitly accounts for the compositional task structure observed in EHR datasets. 
Specifically, we decompose the $Q$-function into one component that represents the value function of the background variant and a separate component for each foreground variant. Each foreground component captures the difference between the value function of the background variant and the corresponding foreground variant. The component capturing the value function for the background variant is easier to learn because there is far more data and treatment responses are more homogeneous across the respective patient sub-population. The reward function is shared across all task variants, but each task can have different state transition probabilities. To infer optimal actions for each of the foreground variants, we use both the background component and a variant-specific foreground component.
By decomposing the $Q$-function in this way, our approach capitalizes on shared structure across task variants while accurately modeling the differences between the background variant and each foreground variant. 
Our results demonstrate that this decomposition outperforms baseline methods for EHR datasets, particularly in cases with substantial group sample imbalance. 

To the best of our knowledge, we are the first to identify and leverage compositional structures in EHR data to enable effective transfer of learning across patient sub-populations. Our contributions to the existing literature that accommodates compositional task structures are as follows:
\begin{enumerate}
    \item We introduce compositional fitted $Q$-iteration (CFQI),
    a specific instance of compositional $Q$-learning, that decomposes the $Q$-value function into a background module and foreground variant-specific modules. 
    \item Unlike related approaches, CFQI does not require sequential ordering of task variants from easiest to hardest, nor does it require estimation of the relationships between the task variants.
    \item CFQI can learn precise policies even in the presence of extreme sample imbalance.
\end{enumerate}
This paper is structured as follows: Section \ref{sec:related} covers related approaches and Section \ref{sec:methods} reviews fundamental topics and formally introduces the method. In Section \ref{sec:results}, we apply CFQI to a simulated Cartpole environment and demonstrate that it outperforms baseline methods on a set of quantitative metrics. We also learn electrolyte repletion policies for renal and non-renal patients in MIMIC-IV, and find that CFQI resembles clinician policies and produces more tailored treatment recommendations for renal patients than baseline methods. Finally, in Section \ref{sec:discussion}, we conclude with a summary of key findings, discuss broader implications of this work, and outline future research directions.

\section{Related Work}
\label{sec:related}
There are several possible strategies to address known group structure in reinforcement learning settings. These range from compositional $Q$-learning,  to Multi-task and Meta-RL algorithms which are broader generalizations. \citet{sdt} first introduces the idea of a sequential decision setting in which tasks are ordered sequentially with simpler sub-tasks preceding more complex ones. All tasks share the same transition dynamics, state-, and action-spaces, but can differ in reward functions. The proposed approach learns a set of $Q$-learning modules, and uses a gating architecture to determine which modules to use to solve each task. Continual RL~\citep{continual_learning} builds on the compositional framework by characterizing a setting in which an agent solves a stream of tasks, where each task is increasingly complex. A continual RL agent is able to effectively retain information from easier tasks, and unlike compositional $Q$-learning, does not require a specific gating architecture. 

Furthermore, ~\citet{Thrun95c} suggests a function approximation approach for effective information reuse in composite tasks, and ~\citet{parr_russell} frames the compositional task setting as a ``hierarchy of machines'', using a modified $Q$-learning approach to speed up decision-making. \citet{khetarpal2022continual} reviews more related works for compositional tasks. $Q$-value decomposition has also been used to speed up value function approximation~\citep{vanseijen2017hybrid} and enable better reward explanations~\citep{Juozapaitis2019ExplainableRL}. In these settings, the reward function is a sum of separate $Q$-function modules. $Q$-value decomposition with factored action spaces has also been shown to improve policy learning when there are few samples~\citep{tang2023leveraging}. 

The compositional task structures explored in related work are similar to our setting, as both involve aggregating knowledge from simpler tasks to aid in solving more challenging ones. An important distinction is that in previous works, sequential decision tasks may have a prescribed order and can be distinct, with separate reward functions. In contrast, our setting is restricted to one task, but with several variants, which all must be solved simultaneously. These variants share the same reward function and state- and action-spaces, but vary in difficulty due to altered transition dynamics and smaller dataset sizes. 

Such a setting can in theory be framed using Multi-task or Meta-RL algorithms, which solve multiple tasks simultaneously while benefiting from shared inter-task structure~\citep{caruana, gupta2020unsupervised}. This structure can be formalized with hidden parameter MDPs (HiP-MDPs)~\citep{doshi2016hidden,killian2017robust, yao2018direct,perez2020generalized}, which parameterize related MDPs using low-dimensional latent factors. However, a key challenge in these approaches is determining how to effectively combine information across tasks. While multi-task and meta-RL methods represent a broader generalization of the compositional setting, the process of learning the relationship between the tasks can be challenging. This can be due to negative interference, where some information is only useful for one task but not for others~\citep{teh2017distral}. Furthermore, choosing which tasks to learn together or how much information can be shared across tasks is an ongoing research problem~\citep{standley2020tasks}.  In contrast, our compositional setting assumes a 
known relationship between task variants, which our approach takes advantage of. 

\section{Methods}
\label{sec:methods}
\subsection{Notation and preliminaries}
A Markov decision process (MDP) is a discrete-time stochastic control process defined by a $5$-tuple $(\mathcal{S}, \mathcal{A}, P, r, \gamma)$, where $\mathcal{S}$ is a continuous state space and for each state $\mathbf{s}\in\mathcal{S}$, $\mathbf{s} \in \mathbb{R}^p$; $\mathcal{A}$ is a discrete set of actions; $P(\mathbf{s}_{t+1} | \mathbf{s}_t, a_t)$ captures state transition probabilities from time $t$ to $t+1$; $r : \mathcal{S} \times \mathcal{A} \rightarrow \mathbb{R}$ is a reward function which we assume to be deterministic; and $\gamma \in [0, 1]$ is a discount factor. A policy $\pi : \mathcal{S} \rightarrow \mathcal{A}$ is a mapping from states to actions that describes an agent's strategy within an MDP. 

The goal of an RL agent is to maximize the value function, 
\begin{equation}
\label{eqn:value_fn}
V^\pi(\mathbf{s}) = \mathbb{E}_{\pi}[\sum_{t=0}^\infty \gamma^t r_t|\mathbf{s}_0=\mathbf{s}].
\end{equation}
The $Q$-function of a policy $\pi$, which represents the expected future accumulative reward $Q$ after taking action $a$ from state $\mathbf{s}$, can be written as:
\begin{equation}
\label{eqn:q_fqi}
Q^\pi(\mathbf{s}_t, a_t) = r(\mathbf{s}_t, a_t) + \gamma \mathbb{E}_{\mathbf{s}_{t+1} \sim P}[V^\pi(\mathbf{s}_{t+1})],
\end{equation}
where the expectation is taken over the possible next states given the transition probabilities $P(s_{t+1}|s_t,a_t)$.
The principle of Bellman optimality~\citep{Bellman_1952} states that a policy $\pi^\star$ is optimal for a given MDP if and only if, for every state $\mathbf{s} \in \mathcal{S}$ and for every action $a \in \mathcal{A}$, it holds that
\begin{align}
    Q^{\pi*}(\mathbf{s}_t,a_t) ={r(\mathbf{s}_t, a_t) +  \gamma \mathbb{E}_{\mathbf{s}_{t+1}\sim P}[\max_a Q^{\pi *}(\mathbf{s}_t, a_t)]}.
\end{align}
\subsection{Fitted Q-iteration (FQI)}
FQI learns the optimal policy by approximating the $Q$-function iteratively in an offline fashion~\citep{ernst2005tree}. 
FQI is appropriate for RL in medical settings because it does not require knowledge of the state transition probabilities, which are unknown or difficult to approximate using EHR. 
FQI first posits a family of functions $\mathcal{Q}(s,a)=\{f(\mathbf{s}, a; \theta), \theta \in \Theta\}$, parameterized by $\theta$, that are meant to approximate the $Q$-function. Commonly used function classes are regression trees~\citep{ernst2005tree} and neural networks~\citep{riedmiller2005neural}. 
It then uses an iterative algorithm that alternates between updating the $Q$-values using the current function approximation and fitting a regression problem that optimizes the parameters $\theta$.
Suppose we have a set of $N$ samples each represented by a 4-tuple $\{(\mathbf{s}^n_t, a^n_t, \mathbf{s}^n_{t+1}, r^n_t)\}_{n=1}^N$, from a policy executed within an MDP, where $t$ is any time step during a trajectory. On iteration $k$ of the algorithm, 
FQI first updates its estimate of the $Q$-function as parameterized by $\theta_{k-1}$ according to the Bellman equation:
\begin{equation}
\label{eqn:q}
    \widehat{Q}_k(\mathbf{s}_t^n, a_t^n) = r_{t}^n + \gamma \max_{a\in \mathcal{A}} f_{\theta_{k-1}}(\mathbf{s}_{t+1}^n, a)
\end{equation}
for $t=1,\dots,T$.
The second step updates $\theta$ by minimizing a loss that measures the discrepancy between the current $Q$-function estimate and the approximating function $f_{\theta_{k-1}}$, i.e.,
\begin{equation}
\label{eqn:theta}
    \theta_k = \arg\min_{\theta} \sum_{n=1}^N \mathcal{L}\left(\widehat{Q}_k(\mathbf{s}^n, a^n), f_{\theta}(\mathbf{s}^n, a^n)\right),
\end{equation}
where we choose $\mathcal{L}$ to be mean squared error loss, but other choices are appropriate as well.  After $K$ iterations, FQI's approximation to the optimal policy is given by $\pi(\mathbf{s}) =  \max_{a\in \mathcal{A}} f_{\theta_K}(\mathbf{s}, a).$ (see Appendix Algorithm \ref{alg:train_fqi} for the full FQI algorithm).
\subsection{Compositional Fitted Q-iteration (CFQI)}
We now introduce CFQI, an extension of FQI that estimates task variant-specific $Q$-functions to solve compositional tasks. Like FQI, CFQI is a framework that can use any suitable function class to approximate the $Q$-function; we describe the general CFQI framework with minimal assumptions about its parameterization.

A compositional task is one composed of one background variant and one or more foreground variants. We assume that we have $Z$ foreground variants in total and that the state- and action-spaces as well as the reward function are shared across all variants. We allow the transition dynamics between the variants to differ; thus the corresponding policies that solve each variant are likely to be distinct.
To account for this task design, we impose a specific compositional structure on the family of approximating functions. 
In CFQI, we define the approximating 
$Q$-function $f$ as a composition of a background variant $Q$-function and a sum of foreground variant-specific residual $Q$-functions,
\begin{equation}
    \label{eq:cfqi_generic}
    f(\mathbf{s}, a, z; \theta) = g_b(\mathbf{s}, a; \theta_b) + 
 g_{f_z}(\mathbf{s},a; \theta_{f_z}),        
\end{equation}
where $\theta = \{\theta_b, \theta_{f_1} \dots \theta_{f_Z}\}$, $g_b$ is a function modeling the $Q$-function of the background variant, and for each foreground variant $i \in [Z]$, 
the corresponding $g_{f_z}$ models the deviation between the background variant and foreground-specific variant. We assume that the group label corresponding to each sample is known. Intuitively, this structure allows us to compile information used to solved the background variant in $g_b$, and learn the residual information needed for each of the foreground variants using the $g_{f_z}$ parameters. Thus, $f$ is a function that approximates the $Q$-function for each of the foreground variants, while $g_b$ approximates the $Q$-function for the background variant.

\subsection{Inference for CFQI}
We use a multi-stage training procedure to fit CFQI. In the first stage, we learn the background function $g_b$ using only background training samples. In the second stage, we freeze the background parameters and train each of the foreground-specific functions using only foreground group-specific samples. Each of the functions uses independent optimizers. We use the Adam optimizer with a higher learning rate for the background variant, and a lower learning rate for each of the foreground variants. We find that this can lead to less variance and faster convergence when incorporating samples from smaller foreground variant groups. 
In practice, we prefer a simpler function class for each $g_{f_z}$ because we often have fewer samples for the minority foreground groups and want to avoid overfitting. 
In order to predict $Q$-values for a background sample, CFQI uses just $g_b$. For a foreground sample from group $z$, CFQI uses both $g_b$ and the foreground group-specific parameters $g_{f_z}$ (see Appendix Algorithm \ref{alg:train} for the full CFQI algorithm).

\section{Experiments and results}
\label{sec:results}
\subsection{Baselines and evaluation metrics}
In our experiments, we compare CFQI to two variations of FQI. The first variation trains FQI using the union of background and foreground samples without any consideration of the group label. Here, a single fitted model captures the optimal policy for both groups.
We also compare against separate FQI policies, which train distinct models to learn policies for the background and foreground groups. We expect separate FQI policies to perform well if there is no class imbalance, and a single FQI policy to perform well if the foreground and background variants are similar.
For context, we also compare our approach to an agent that acts randomly, which we expect will perform the worst. For the MIMIC-IV experiments, we also compare CFQI to a clinician, or oracle policy, that takes the actions administered in the dataset. 
While we recognize that multi-task and meta-RL algorithms can be used here, we instead opt for baselines that also assume knowledge of the compositional task structure. Specifically, these baselines assume that all task variants have the same objective and share the same reward function.

We measure performance in the Cartpole task by counting the number of timesteps the pole stays upright. For electrolyte repletion, we measure performance qualitatively using the log of the value of the policy (Equation \ref{eqn:value_fn}). 

\subsection{OpenAI Gym Cartpole}
We first validate the performance of CFQI using the Cartpole environment~\citep{brockman2016openai}, which consists of an unstable pole attached to the top of a cart. The agent's objective is to keep the pole balanced upright for $1,000$ steps; this can be done by strategically applying a force on the cart to the left or right at each time step (see Appendix \ref{appendix:cartpole} for MDP). 

We adapt the Cartpole environment to accommodate a compositional setting in which there is one background and one foreground variant. The background variant is the original Cartpole, and the foreground variant includes a constant force of $c$ Newtons that pushes the cart to the left. Unless otherwise specified, $c=5$. Although the objectives for the two task variants are identical, an agent will require a different policy to succeed in each of the variants; the foreground variant requires a policy that counteracts the constant leftward force on the cart.

\textbf{With equal dataset sizes, CFQI keeps the pole upright longer than the baselines.} 
Our results demonstrate that CFQI gathers the highest reward in comparison to the baselines. Its performance is comparable to separate FQI policies trained on either task variant
(Appendix\ref{apd:equal_datastes}). 
We hypothesize that an FQI policy trained using the union of the two datasets cannot capture the optimal policy for each task variant because samples from each dataset provide conflicting information when aggregated to learn a policy.

\textbf{When the force modifying the foreground variant increases, CFQI maintains performance in the background and degrades more slowly in the foreground than the baselines.} Here, we progressively increase the magnitude of the force pushing the cart left to mimic a scenario in which the foreground variant is increasingly distinct from the background variant. We find that CFQI maintains high performance in the background variant, and degrades more slowly than the baselines in the foreground variant as the force increases (Appendix \ref{sec:cartpole_c_increases}). This also shows that there are limits to the datasets in which CFQI is applicable; if the foreground variant is substantially distinct from the background variant, the performance degrades. 

\begin{figure*}[htbp]
  {\begin{minipage}[t]{0.5\linewidth}
      {\includegraphics[width=\linewidth]{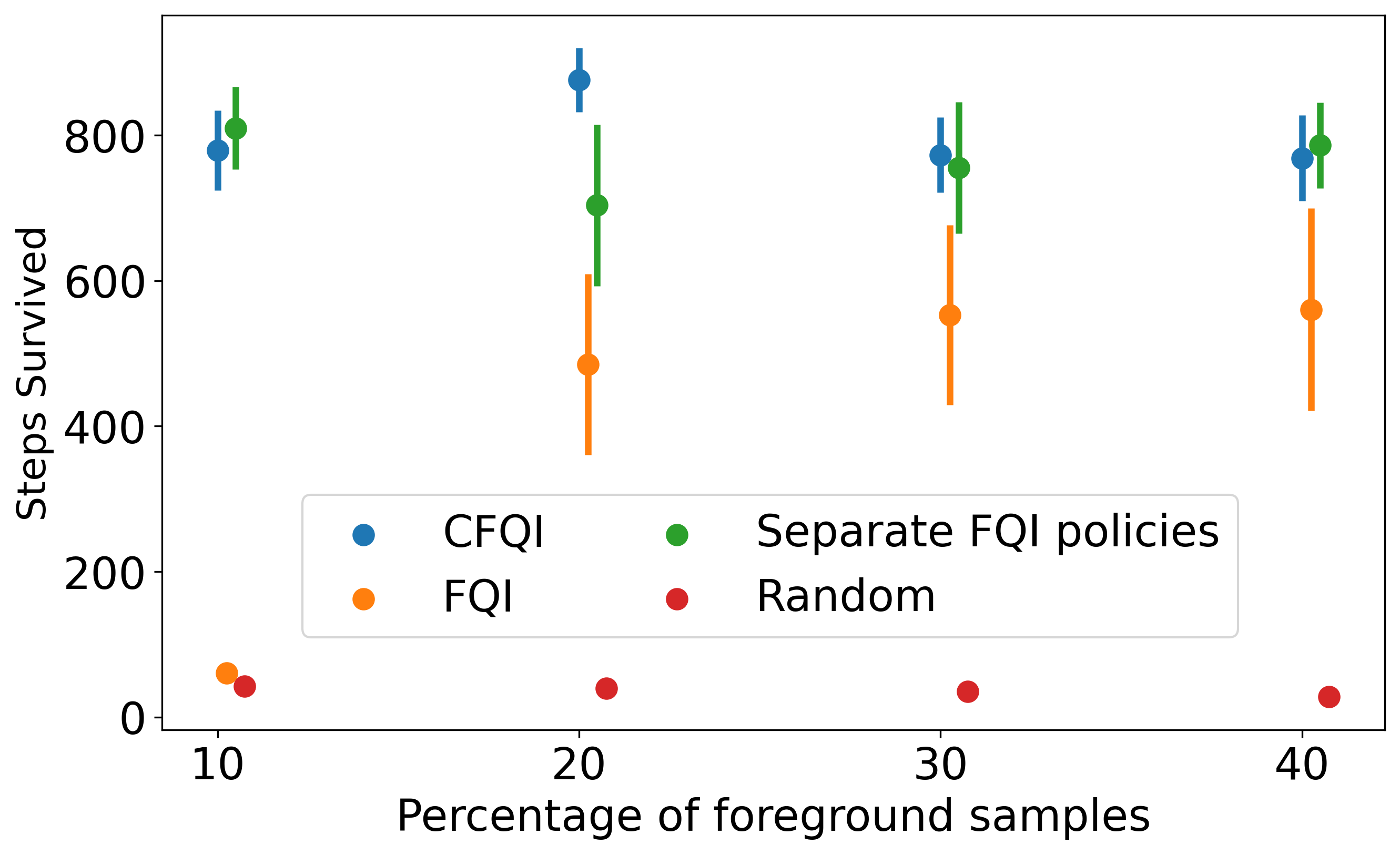}}
  \end{minipage}}
  \hfill
  {\begin{minipage}[t]{0.5\linewidth}
      {\includegraphics[width=\linewidth]{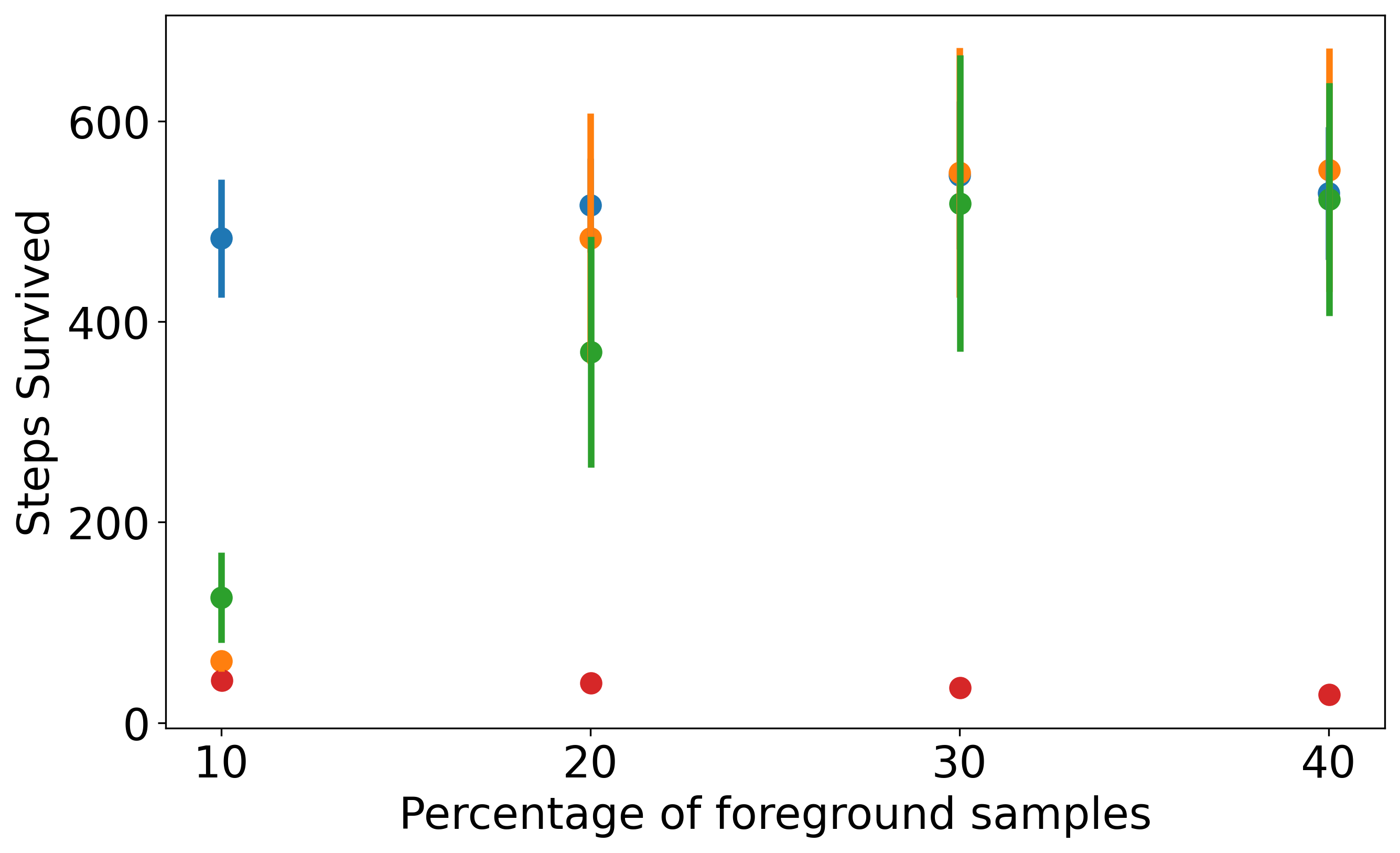}}
  \end{minipage}}
  \caption{\textbf{CFQI is robust to class imbalance.} (Left) Background performance; (Right) Foreground performance. The total number of training samples is $400$, and the fraction of samples (x-axis) in the foreground data varies, where $50\%$ corresponds to balanced sizes. The y-axis reports the number of steps the pole stays up (higher is better).
  Error bars represent standard error across 50 evaluations. 
  CFQI maintains superior or comparable performance to the baselines despite extreme class imbalance.}
  \label{fig:class_imbalance}
\end{figure*}

\textbf{CFQI outperforms baselines even with severe class imbalance.} 
In medical settings, we expect that we will have fewer samples for the foreground variant than the background one, and thus the datasets for compositional tasks have imbalanced sizes.
CFQI can adapt to a wide range of class imbalance; in the most extreme imbalance we tested, CFQI vastly outperforms the baselines (Figure \ref{fig:class_imbalance} and Appendix \ref{sec:imbalance}). 

\subsection{Treating potassium repletion in renal and non-renal patients}
Next, we evaluate CFQI in a medical setting using EHRs from the Medical Information Mart for Intensive Care (MIMIC-IV) series~\citep{mimicivdataset}. Our goal is to identify a policy for maintaining healthy levels of electrolytes in patients. Determining when to replete electrolytes is a difficult problem and over-repletion has been linked to patient mortality and complications post-surgery~\citep{voldby_fluid_2016}. We use CFQI to model potassium repletion strategies for renal patients (foreground variant) and non-renal patients (background variant). We use a patient cohort in which every patient receives potassium repletion; some have end-stage renal disease (ESRD)($n = 2,258$) and others have functioning kidneys $(n = 36,023)$. Pre-processing details are described in Appendix \ref{appendix:mimic}.  

Now, we use CFQI to identify treatment policies for the two patient groups. Our goal is to determine the dosage of IV potassium to administer to a patient across the duration of their hospital stay. We discretize the action space into seven levels of repletion: \emph{no repletion} (0 mEq), 3 stages of \emph{low repletion} (10, 20, 30 mEq), and 3 stages of \emph{high repletion} (40, 50, 60 mEq). 
The reward function takes as input both the patient state and repletion action, and produces higher reward when a patient's measured potassium is within the reference range (3.5-4.5 mmol/L), and lower reward if the potassium is outside of this range. It also considers the cost of repletion~\citep{Prasad_2020} (see Appendix \ref{appendix:mimic} for further training details). 
We report results for CFQI and FQI using an importance sampling-based logged estimate of the value of the learned policy, 
$log(V^{\pi})$~\citep{Gottesman2018EvaluatingRL}(Table \ref{tab:mimic_table}; Equation \ref{eqn:value_fn}). 
\begin{table}[htb]
\resizebox{\columnwidth}{!}{%
\begin{tabular}{|c|c|c|}
\hline
    & \text{Background}        & \text{Foreground}        \\
    \hline
Clinician (Expert) & 1.40 $\pm$ 0.56 & 10.57 $\pm$ 0.96  \\ \hline
CFQI & \textbf{-0.97 $\pm$ 1.84} & \textbf{9.18 $\pm$ 1.7} \\
\hline
FQI & -4.38 $\pm$ 0.12 &  -6.13 $\pm$ 0.129 \\
\hline
Separate FQI & \textbf{-0.652 $\pm$ 1.84}  & -27.95 $\pm$ 0.46 \\
\hline
Random &  -6.49 $\pm$ 2.1 &  -28.4 $\pm$ 1.98 \\
\hline
\end{tabular}
}
\caption{\textbf{Evaluation of policies using $log(V^{\pi})$.} Results are reported across 10 train/test splits as $m \pm se$, where $m$ is the mean and $se$ is standard error. CFQI performs most comparably to the clinician policy, 
and a separate FQI policy suffers for renal patients, due to the small sample size. A single FQI policy also falters, likely due to naive aggregation of information contained in the task-variant datasets.}
\label{tab:mimic_table}
\end{table}

\textbf{The value of the CFQI policies is comparable to the value of an expert clinician policy.} The value of the policy corresponds to the total reward received by an agent that performs according to the policy. A higher value indicates a better policy. CFQI policies outperform all baselines particularly for renal patients, even with the extreme class imbalance present (Table~\ref{tab:mimic_table}). More importantly, the values of CFQI policies approach that of the clinician policy for both renal and non-renal patients, meaning that they perform nearly as well as clinicians. For non-renal patients (background variant), the separate FQI policy performs comparably to the expert policy, likely due to the large sample size. In contrast, the eparate FQI policies fails for renal patients in part because of the corresponding small sample size. We hypothesize that a single FQI policy naively aggregates information across the background and foreground datasets, even though samples from both groups may provide conflicting information. 

\begin{figure*}[htbp]
  {\begin{minipage}[t]{0.45\linewidth}
      {\includegraphics[width=\linewidth]{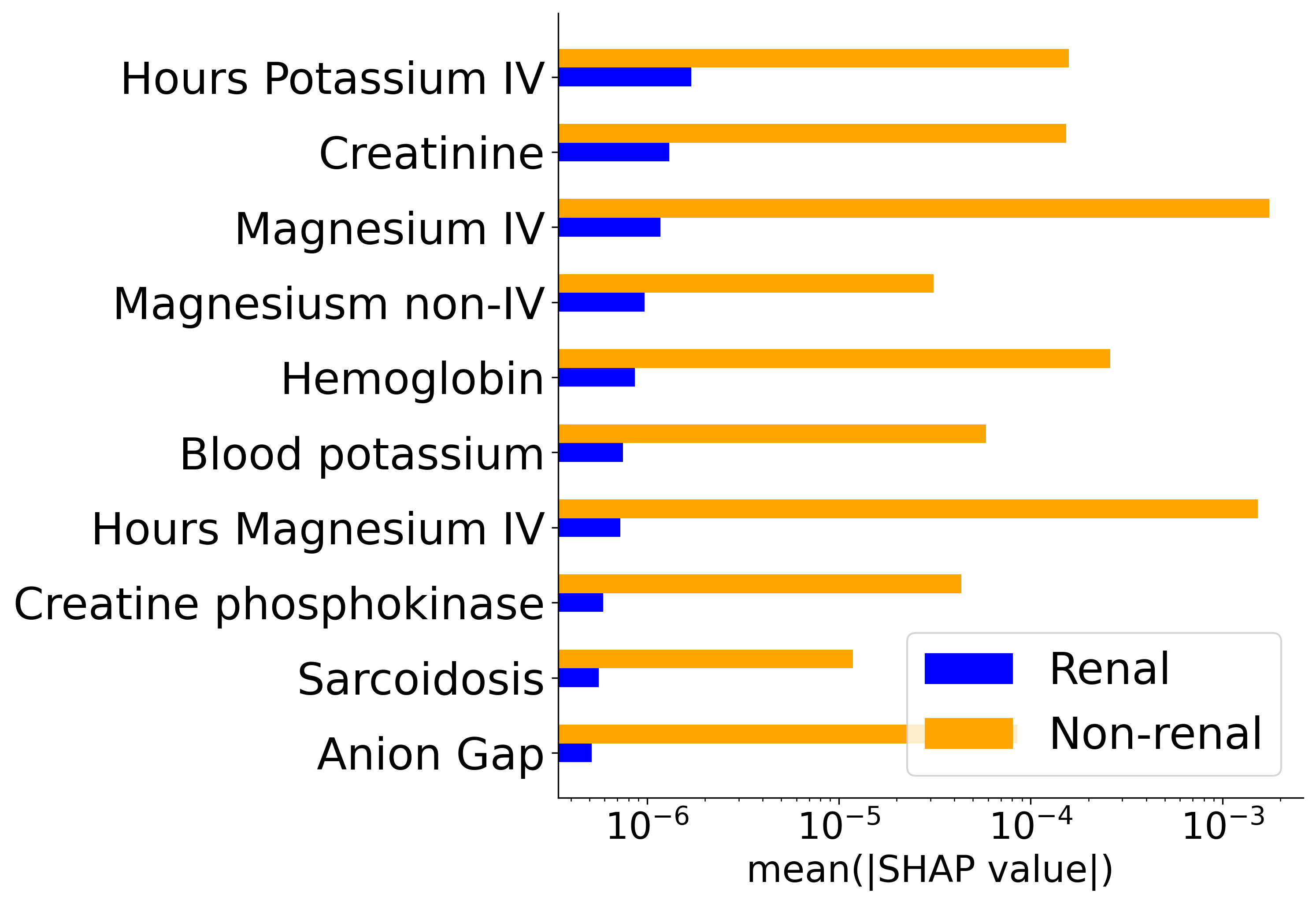}}
  \end{minipage}}
  \hfill
  {\begin{minipage}[t]{0.45\linewidth}
      {\includegraphics[width=\linewidth]{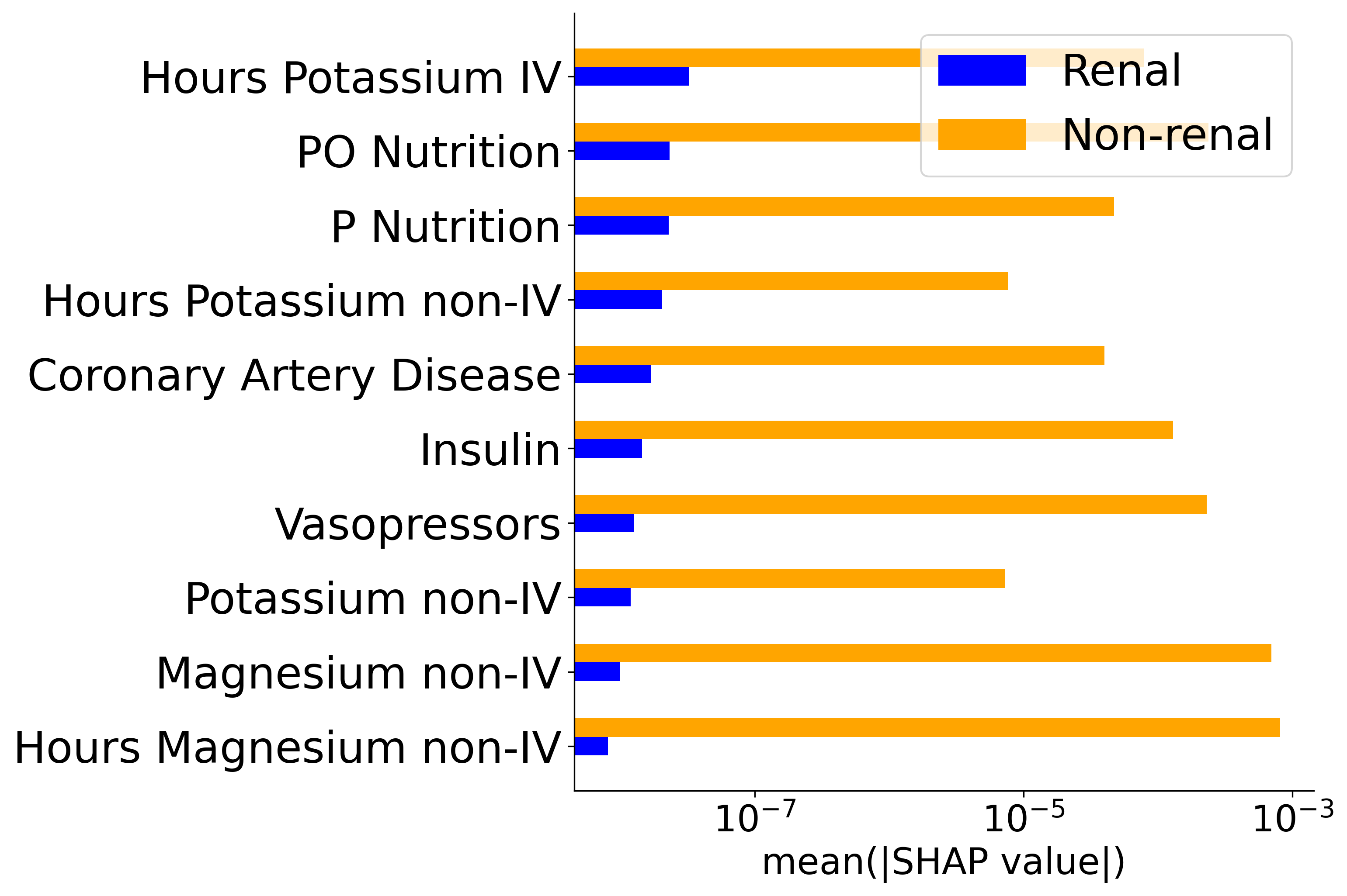}}
  \end{minipage}}
  \caption{\textbf{CFQI relies on more relevant state features than FQI.} 
  The x-axis represents mean Shapley value and the y-axis shows the features used for prediction. 
  (Left) CFQI primarily uses existing potassium treatment and creatinine levels for prediction. 
  (Right) FQI does not use creatinine to predict treatment for renal patients, and relies on existing electrolyte treatments for both groups. CFQI appears to use creatinine to distinguish between groups.
  The magnitude of Shapley values differs between renal and non-renal patients due to sample imbalance.}
  \label{fig:mimic_shap}
\end{figure*}

\textbf{CFQI relies on relevant state features to learn electrolyte repletion policies.} To understand CFQI's performance, we study feature importance metrics using Shapley values~\citep{shap}. Higher Shapley values for a feature indicate that the output is more closely tied to fluctuations in that feature. CFQI most relies on existing electrolyte treatments and blood creatinine to make predictions (Figure \ref{fig:mimic_shap}). Higher levels of creatinine indicate worse kidney function~\citep{creatinine}, and high creatinine is a characteristic of renal disease. In contrast, FQI primarily uses pre-existing potassium and nutrition treatment, but does not use creatinine to distinguish the patient populations. 
\begin{figure*}[htbp]
  {
    \begin{minipage}[t]{\linewidth}
        {\includegraphics[width=0.3\linewidth]{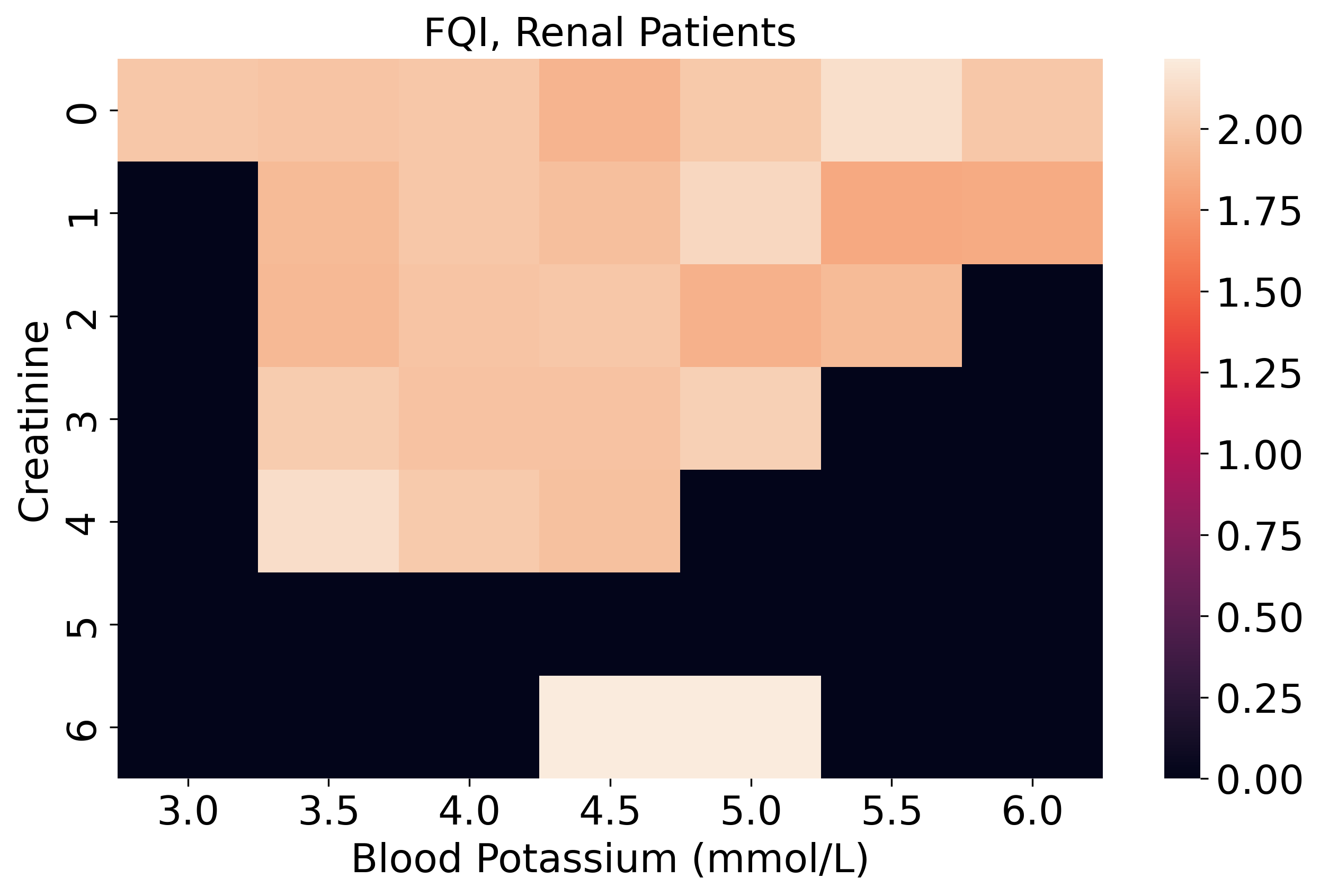}}
        {\includegraphics[width=0.3\linewidth]{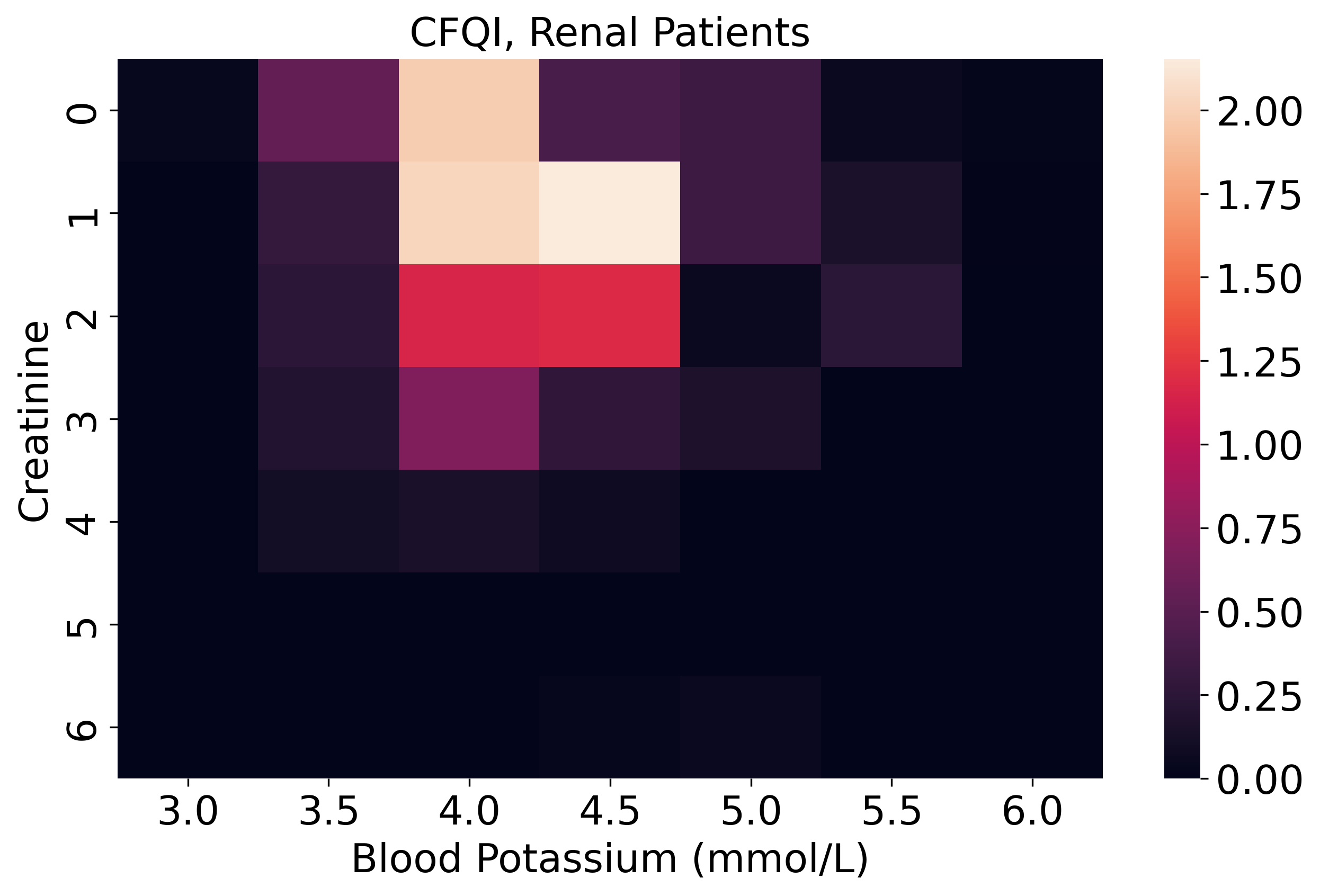}}
        {\includegraphics[width=0.3\linewidth]{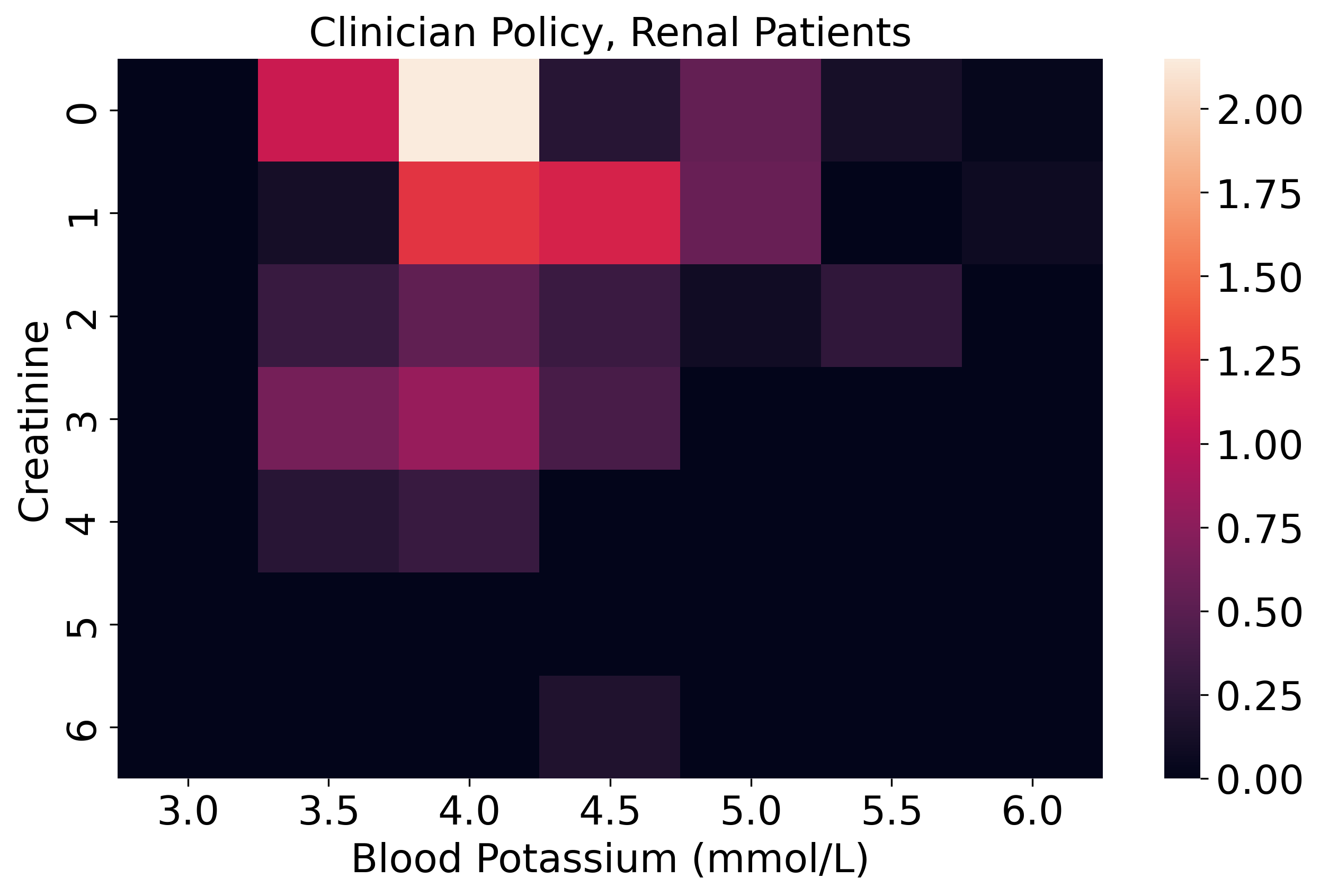}}
    \end{minipage}
    \begin{minipage}[t]{\linewidth}
        {\includegraphics[width=0.3\linewidth]{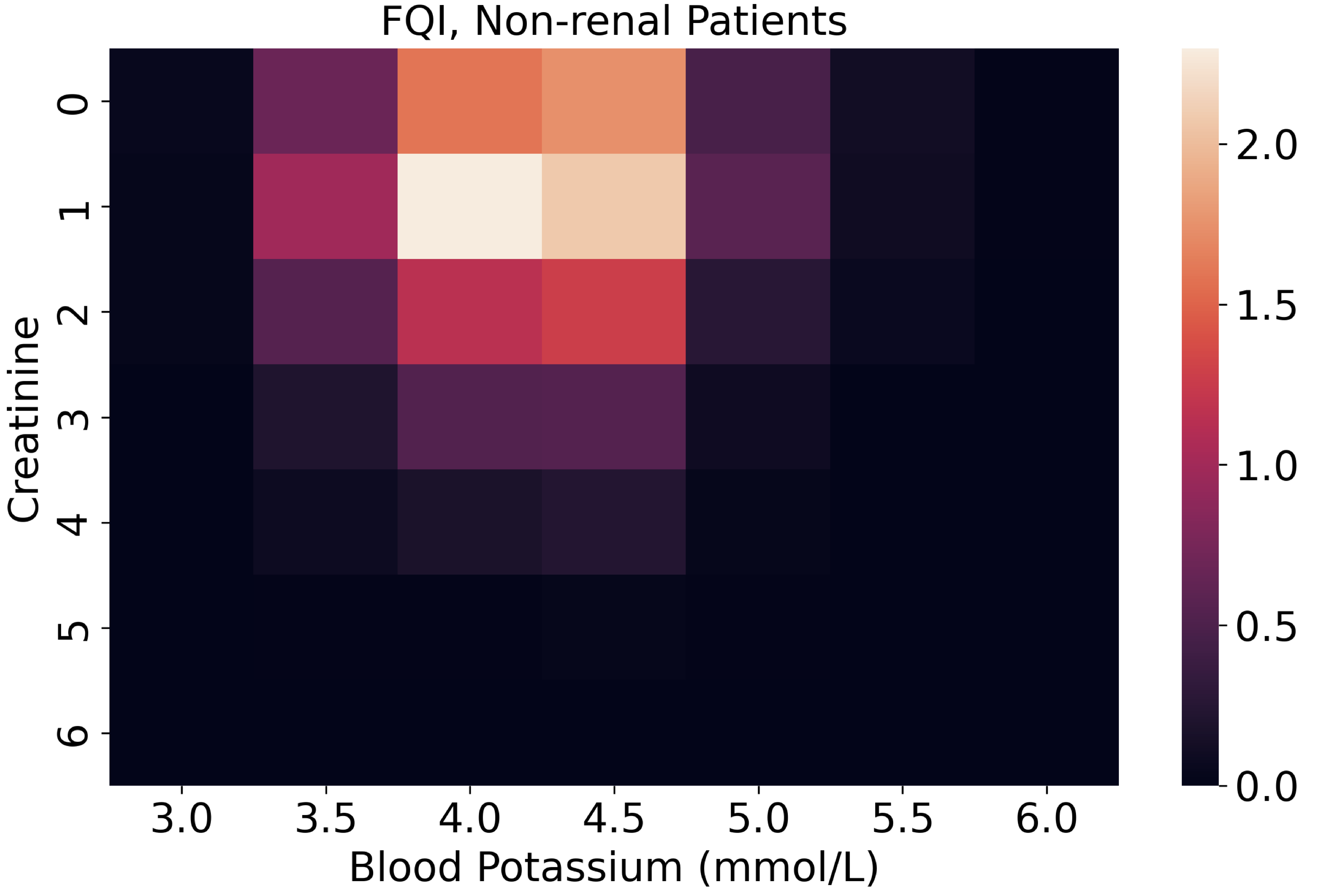}}
        {\includegraphics[width=0.3\linewidth]{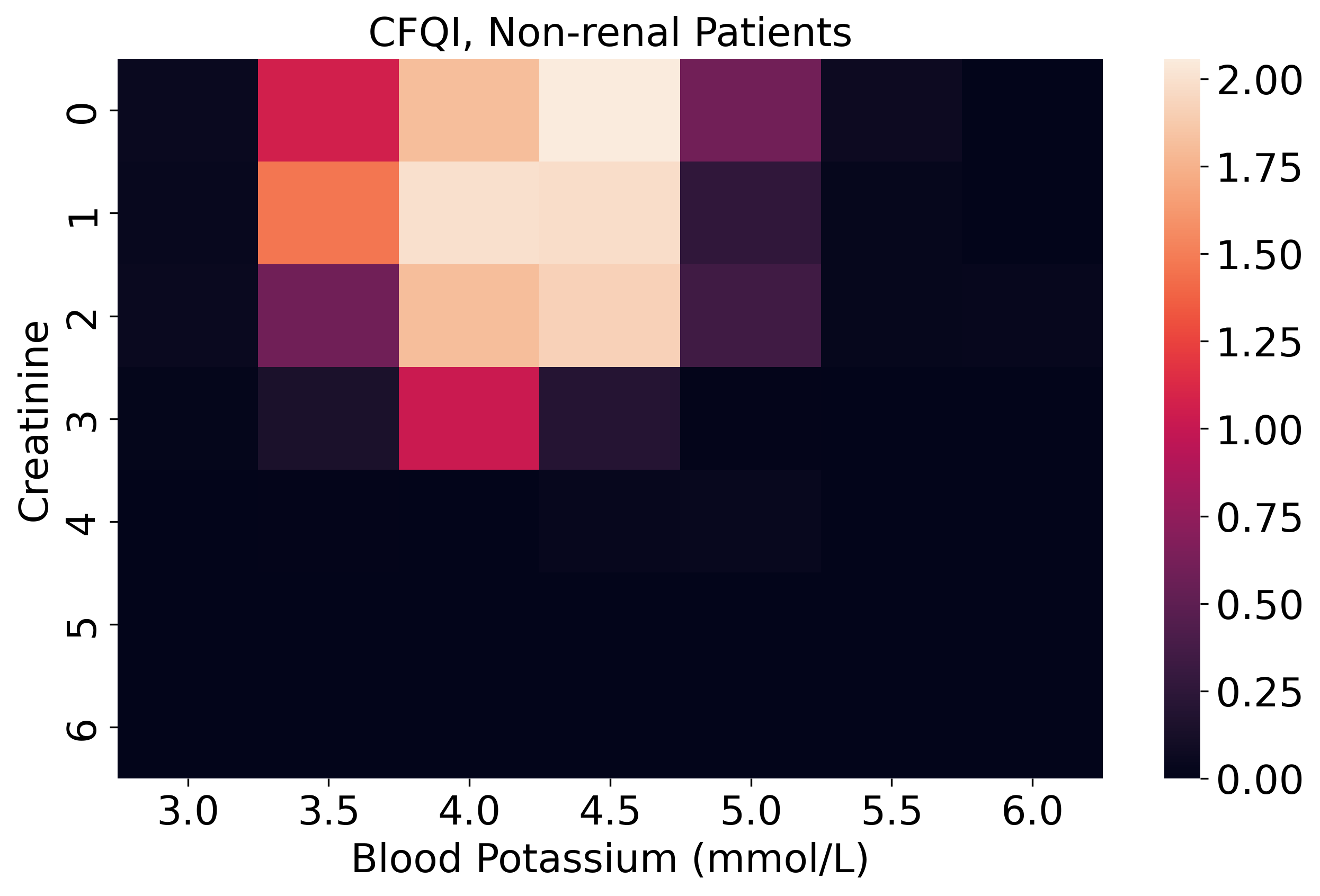}}
        {\includegraphics[width=0.3\linewidth]{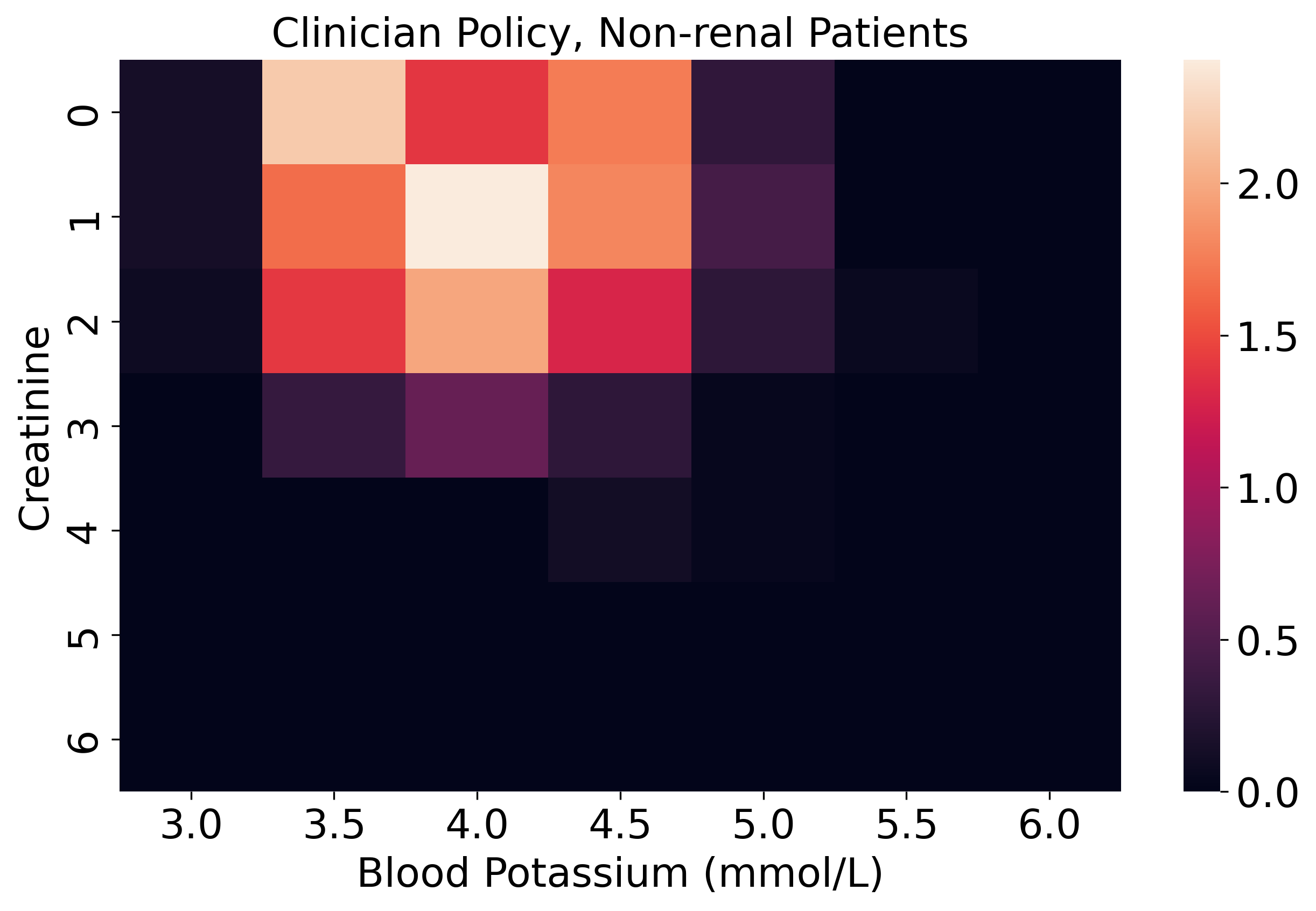}}
    \end{minipage}
  }
   \caption{\textbf{Visualizing CFQI and FQI policies for renal and non-renal patients.}
   The x-axis represents blood potassium level and y-axis represents blood creatinine level. Each heatmap shows the policy's repletion recommendation discretized into three levels of repletion (no repletion = 0, low repletion = 1.0, high repletion = 2.0); the lighter the color, the higher the repletion level. The first row corresponds to renal patients, and the second row to non-renal patients.
   While the FQI policy is similar to the clinician policy for non-renal patients, CFQI more resembles the clinician policy for both groups, suggesting that CFQI can better recommend repletion. 
   }
   \label{fig:heatmaps}
\end{figure*}

\textbf{When visualizing treatment recommendations across creatinine and blood potassium levels, CFQI aligns more closely with the clinician policy than FQI.} We create synthetic patient states by densely sampling a grid with measured potassium and creatinine levels, while holding all other features fixed. We expect blood potassium and creatinine to have the largest effect on treatment recommendations; treatment should be largely restricted to lower blood potassium and creatinine levels. Given a patient's state features, we use the trained $Q$-value functions to select the optimal action (Figure \ref{fig:heatmaps}). For simplicity, we use three classes of repletion: no repletion, low repletion, and high repletion. CFQI recommends less aggressive repletion for renal patients than non-renal patients, especially when the patient's potassium is below the reference range or when creatinine is high. Meanwhile, FQI recommends more uniform treatment across potassium and creatinine levels for renal patients, and treatment only for non-renal patients when potassium and creatinine are low. These results suggest that CFQI policies align better with the existing medical practice for both groups~\citep{creatinine}.

\section{Discussion}
\label{sec:discussion}
We introduce compositional fitted $Q$-iteration (CFQI), to design RL policies for EHR datasets with known compositional task structure. Related methods for learning these policies either require large datasets or impose restrictive assumptions on the dataset structure. Our results demonstrate that CFQI outperforms FQI and related approaches in both a 2-group and multi-group setting (Appendix \ref{apd:mgcfqi}). When applied to Cartpole data and a cohort extracted from MIMIC-IV, we find that these policies perform well across all task variants, are robust to sample size imbalance, and only degrade in quality when the difference between task variants is sufficiently large. Furthermore, CFQI policies align with existing medical practice because they resemble the clinician policies, particularly for renal patients.
CFQI is a step towards designing optimal RL policies in datasets with a known compositional structure.

This paper poses several directions for future work. First, other function classes for modeling the compositional $Q$-value function could be considered, such as Gaussian processes, random forests, and other neural network architectures. This may improve the fit to larger datasets by increasing the flexibility of the $Q$-value approximating function. Additionally, we could change the structure of the approximating function $f$ to incorporate other relationships between the $Q$-function of the background variant, $g_b$, and the $Q$-function of the foreground variant $g_{f_z}$. Finally, our work notes that the difference between the task variants affects the quality of the policy, and further characterizing this behavior is important. 

\acks{The authors would like to thank Matthew J{\"o}rke, Alex Nam, Archit Verma and Dylan Cable for feedback. AM was supported in part by a Stanford Engineering fellowship. AM, JY, and BEE are funded in part by Helmsley Trust grant AWD1006624, NIH NCI 5U2CCA233195, and CZI. BEE is a CIFAR Fellow in the Multiscale Human Program.}

\bibliography{ref}
\newpage
\appendix
\section{Code}
\label{appendix:code}
Before CFQI can be deployed in a particular application area, its hyperparameters and architecture should be tuned. All of our experiments were run on an internally-hosted cluster using a 320 NVIDIA P100 GPU whose processor core has 16 GB of memory hosted. Our experiments used a total of approximately 100 hours of compute time. The code for our Cartpole experiments is based on a prior implementation of FQI\footnote{https://github.com/seungjaeryanlee/implementations-nfq}. We implement our models in PyTorch~\citep{paszke2019pytorch} and use a stochastic gradient descent-based optimizer~\citep{Saad_1998}. For all experiments, we use 80\% of our data to train and 20\% of our data to test. Our code is available on \href{https://github.com/bee-hive/cfqi\_public}{Github}. 

\section{Cartpole environment}
\label{appendix:cartpole}
The Cartpole environment contains a cart with an inverted pendulum pole on top. Here, the agent's goal is to move the cart left or right such that the pole remains upright and the cart stays within the bounds of the window box. More formally, we can describe the Cartpole environment as a Markov decision process (MDP) represented by a tuple $(\mathcal{S}, \mathcal{A}, P, R)$. Here, $\mathcal{S} = [x, \dot{x}, \theta, \dot{\theta}]^\top$ is a four-dimensional state space, where $x$ and $\dot{x}$ are the cart's position and velocity, and $\theta$ and $\dot{\theta}$ are the pole's angle and angular velocity; $\mathcal{A} = \{a_\ell, a_r\}$ is the action space, containing actions that apply a force of $10$ Newtons to the cart from the left or right; $P$ is a deterministic function determining state dynamics (based on simple Newtonian mechanics in this case); and $R$ is a reward function, which returns $-1$ when the pole falls, or the cart exits the frame (at which point the episode is terminated) and returns $1$ otherwise. In this work, we adjust the dynamics described by $P$ to simulate different dynamics for the foreground and background environments. Specifically, in the foreground environment, we include a constant leftward force of $c$ Newtons on the cart. In this case, choosing the action ``push left'' results in a leftward force of $10 + c$ Newtons, while the action ``push right'' results in a rightward force of $10 - c$ Newtons. This asymmetry is the driving difference between the background and foreground environments. As such, each environment has the same state space, action space, and reward function but a different state transition function. 

To construct our datasets, we sample Cartpole episodes from the foreground and background variants. Each trajectory starts in a random state and chooses actions uniformly at random until the pole falls over. We evaluate the performance of each policy by observing the duration of time that the pole stays up.

We measure convergence in the Cartpole environment by  evaluating the network on one test sample for every training sample in the batch; if the previous three evaluations were successful (i.e., the pole in the Cartpole environment stayed up for 1000 steps), we deem the parameters converged. 
\section{Algorithms}
\label{appendix:algorithms}
\begin{algorithm2e}[htbp]
\caption{Compositional Fitted Q-iteration (CFQI)}
\raggedright
\label{alg:train}
\DontPrintSemicolon
\vspace{0.33em}
\Indentp{0.4em}
\KwIn{dataset of $N$ $4$-tuples $\{(\mathbf{s}^n_t, a^n_t, \mathbf{s}^n_{t+1}, r^n_t)\}_{n=1}^N$, learning rate} 
\vspace{0.33em}
Randomly initialize background and foreground-specific model parameters $\theta_b, \theta_{f_i}$ for $i \in Z$\;
\While{loss $\mathcal{L}_b$ not converged}{
    \For{$n=1,\dots,N$}{
    \begin{itemize} 
    \item If n is a background sample, compute $\mathcal{L}_b$\;
    \item Compute Q according to Equation \ref{eqn:q}
    \item Update $\theta_b$ according to Equation \ref{eqn:theta} \; 
    \item $\pi_b(\mathbf{s}) \gets \argmax_{a \in A} g_b(\mathbf{s}, a)$ \;
            \Indp \tcp{background policy}
    \end{itemize}
    }
}
\For{$i=1, \dots, Z$}{
    \While{loss $\mathcal{L}_i$ not converged}{
    \For{$n=1,\dots, N$}{
        \begin{itemize} 
        \item If n is a sample from foreground \;
            \Indp  group $i$, compute $\mathcal{L}_i$\;
        \item Compute Q using Equation \ref{eqn:q}
        \item Update $\theta_{f_i}$ using Equation \ref{eqn:theta}\;
        \item $\pi_{f_i}(\mathbf{s}) \gets \argmax_{a \in A} f(s,a, i; \theta_{f_i})$ \;
            \Indp as defined in Equation \ref{eq:cfqi_generic} \;
            \Indp \tcp{foreground policy}\;
        \end{itemize}
    }
    }
}
\Return{$\pi_b, \pi_{f_i}$ for $i\in Z$}
\end{algorithm2e}
\begin{algorithm2e}[t!]
\caption{Fitted Q-iteration (FQI)}
\raggedright
\label{alg:train_fqi}
\DontPrintSemicolon
\vspace{0.33em}
\Indentp{0.4em}
\KwIn{dataset of $4$-tuples $\{(\mathbf{s}^n_t, a^n_t, \mathbf{s}^n_{t+1}, r^n_t)\}_{n=1}^N$, learning rate}
\vspace{0.33em}
Randomly initialize model parameters $\theta$\;
\While{loss $\mathcal{L}$ not converged}{
    \For{$n=1,\dots,N$}{
    \begin{itemize} 
    \item Compute $\mathcal{L}$\;
    \item Compute Q according to Equation \ref{eqn:q}\;
    \item Update $\theta$ using Equation \ref{eqn:theta}\;
    \item $\pi(\mathbf{s}) \gets \argmax_{a \in \mathcal{A}} f(\mathbf{s}, a)$\\ \tcp{policy for all samples}
    \end{itemize}
    }
}
\Return{$\pi$}
\end{algorithm2e}
We use the same number of parameters for CFQI, FQI, and Separate FQI policies, though the network architectures differ between CFQI and FQI variants. With the exception of the imbalanced dataset experiment, all algorithms are trained using the same number of training and test samples. To perform inference using CFQI, we input a state-action pair and the group label into the network. Depending on the group label, the network uses either the background parameters or both the background parameters and one set of foreground parameters to estimate a $Q$-value. To train FQI jointly between the two datasets, the network uses all of the parameters to estimate a $Q$-value for each training sample. 
\section{Cartpole Results}
\subsection{Equal dataset sizes, c=5}
Here we evaluate the performance of CFQI in comparison to the baselines when the force modifying the foreground variant is constant. Our results indicate that CFQI is able to achieve comparable reward to policies that are trained separately on each of the task variants.  (Table \ref{tab:cartpole_constant_c}). 
\label{apd:equal_datastes}
\begin{table}[htb]
\resizebox{\columnwidth}{!}{%
\begin{tabular}{|c|c|c|}
\hline
  & \text{Background}        & \text{Foreground}        \\
    \hline
CFQI  & \textbf{787.93 $\pm$ 100.1} & \textbf{530.8 $\pm$ 114.5} \\
\hline
FQI  & 540.53 $\pm$ 124.8 & 431.26 $\pm$ 120.7  \\
\hline
Separate FQI policies &  785.69 $\pm$ 108.6 & 523.7 $\pm$ 47.4 \\
\hline
Random policy & 42.3 $\pm$ 2.01 &  31.06 $\pm$ 1.09\\
\hline
\end{tabular}
}
\caption{Cartpole performance when $c=5$. We report mean and standard error over 50 evaluations across 10 models for the background and foreground variants. We compare the performance of CFQI to the baselines and find that CFQI can keep the pole upright for at least the same number if not more steps than the baselines. The performance of CFQI is comparable to separate FQI policies trained on each of the task variants separately.}
\label{tab:cartpole_constant_c}
\end{table}
\subsection{Equal dataset sizes, c increases}
\label{sec:cartpole_c_increases}
\begin{figure*}[htbp]
  {\begin{minipage}[t]{0.5\linewidth}
      {\includegraphics[width=\linewidth]{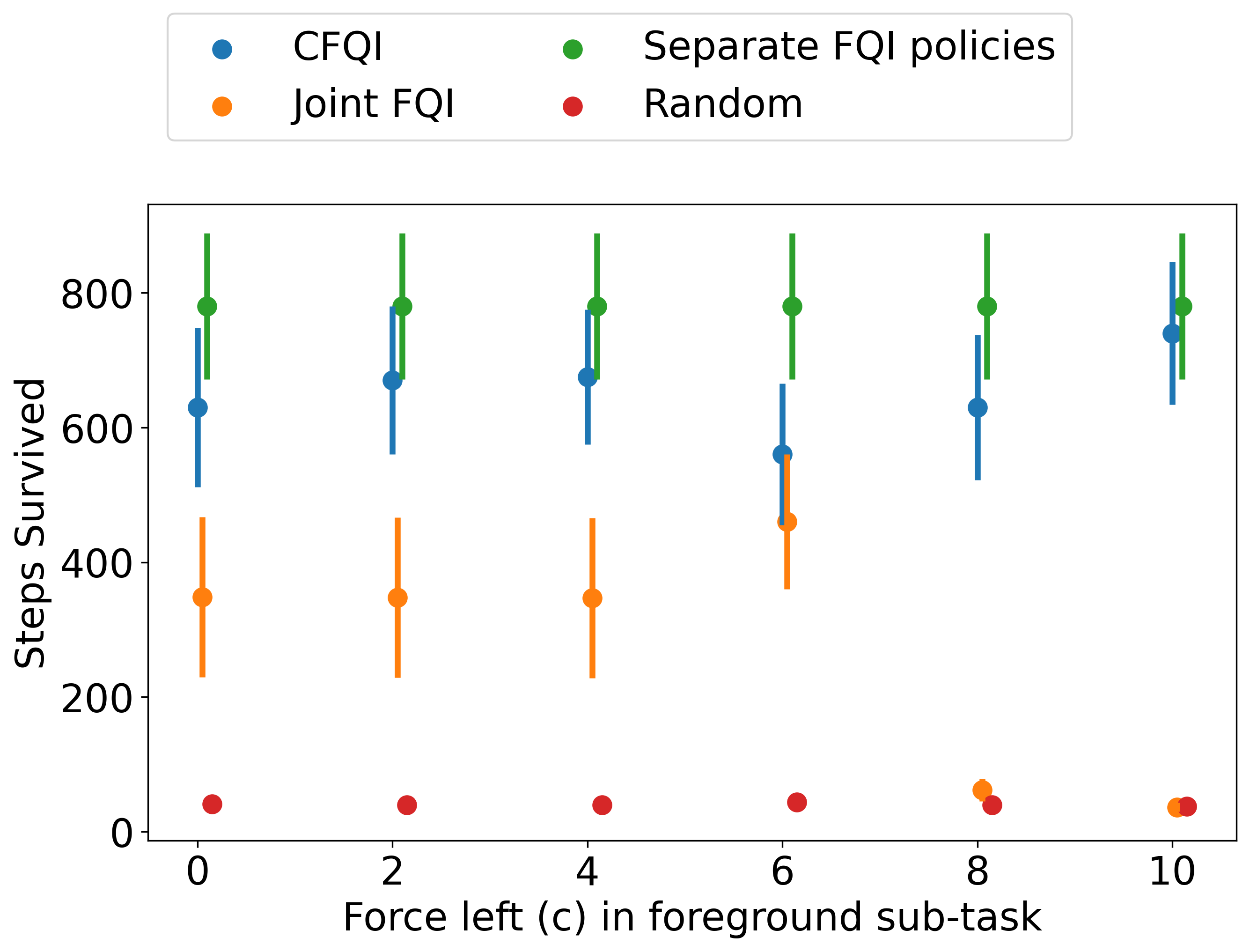}}
  \end{minipage}}
  \hfill
  {\begin{minipage}[t]{0.5\linewidth}
      {\includegraphics[width=\linewidth]{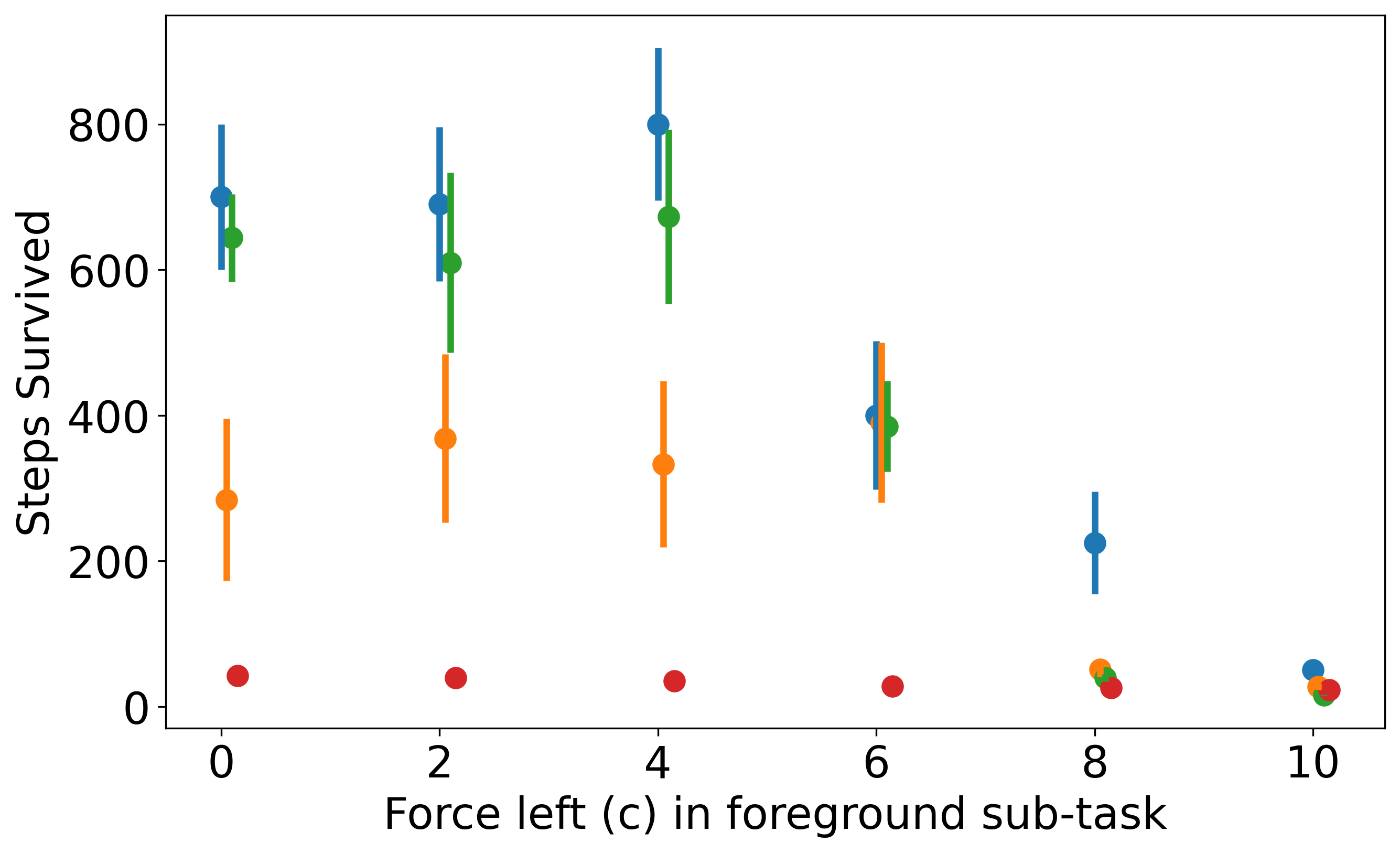}}
  \end{minipage}}
  \caption{As the difference between the background and foreground variants becomes larger, CFQI maintains high performance in the background variant (left) and degrades in the foreground (right) more slowly than related approaches. The x-axis refers to the force modifying the cart in the foreground variant; higher implies that the background and foreground state transition dynamics are more distinct. The y-axis is the number of steps the pole stays up; higher values are more desirable. Error bars indicate standard error across 10 models and 50 evaluations.  }
  \label{fig:force_left}
\end{figure*}
Now, we evaluate CFQI in a setting where the force modifying the foreground Cartpole is not constant. To evaluate the algorithms, we create 6 variations of the Cartpole environment, each corresponding to a different value of $c$, where $c \in \{0, 2, 4, 6, 8, 10\}$. We run $15$ evaluations for each $c$ and each task variant. and find that CFQI outperforms FQI, especially as the foreground and background variants become more distinct (Figure \ref{fig:force_left}). When the foreground and background variants are similar ($c\in [0, 2]$), CFQI and FQI perform comparably, indicating that there is less need to account for group structure. However, as the environments' dynamics become more distinct (as $c$ increases), the need to account for group structure becomes more relevant. CFQI learns policies that maintain high performance in the background, and the performance degrades more slowly in the foreground compared to the baselines. In comparison, separate FQI policies maintain reasonable performance for the background variant but degrade more quickly in the foreground variant. These results also suggest that CFQI can perform well when the background and foreground environments are distinct, but not too distinct; if they differ by too much, the performance of CFQI degrades to that of the baselines. 

\subsection{Dataset sizes vary, c=5}
\label{sec:imbalance}
In many clinical settings, we have access to far less data in the foreground conditions than in the background. Recall that in the electrolyte repletion example introduced in Section \ref{sec:intro}, renal patients, which comprise the foreground variant, make up only about 6\% of the patient cohort. Thus, it is important that CFQI is resilient to sample imbalance and can leverage shared structure in the datasets to estimate foreground-specific structure.

Going back to the setting in which the force modifying the foreground Cartpole is constant ($c=5$), we now evaluate CFQI with imbalanced group sizes. The total number of training samples between the two groups is fixed at $400$, and we set the fraction of samples that come from the foreground data to $10\%, 20\%, 30\%, 40\%,$ and $50\%$ (where $50\%$ corresponds to balanced sizes). We choose these fractions because they reflect percentages of imbalance that we are likely to see in medical data.

Our results indicate that across the range of data imbalance, the policies found by CFQI have equal or better performance in both variants than the baselines (Figure \ref{fig:class_imbalance}). We hypothesize that this is because the foreground policy builds on the background policy and re-uses information learned using the background dataset. This increases effective sample size, and allows CFQI to perform well even with extreme data imbalance. Alternatively, the performance of baseline algorithms in the foreground variant falters, especially when the dataset sizes are extremely imbalanced. This result implies that CFQI is can model foreground variants even when we have access to a small proportion of foreground samples relative to background samples. 

\subsection{Multi-group compositional settings}
\label{apd:mgcfqi}
\begin{figure}[ht!]
\includegraphics[width=\linewidth]{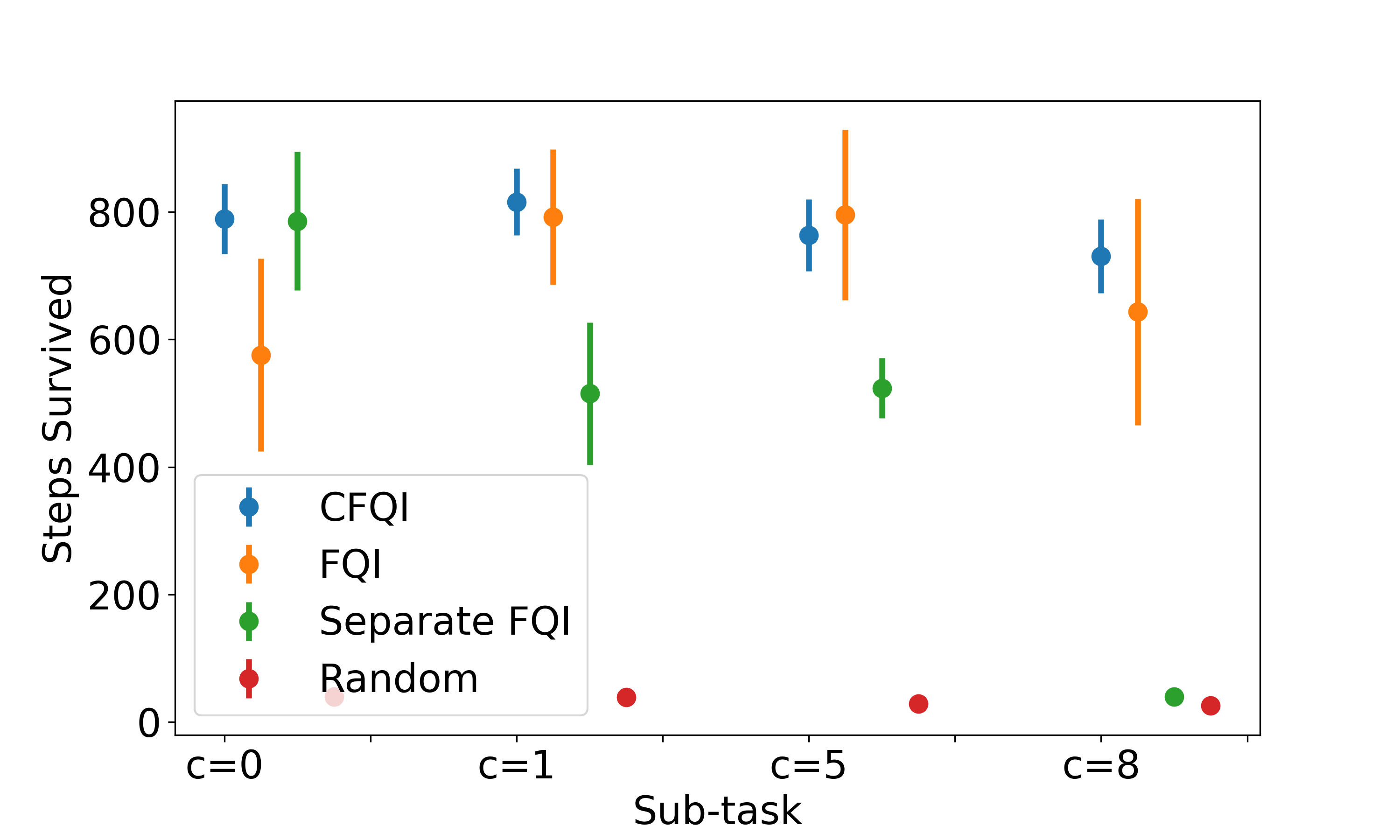}
  \caption{Here we visualize performance across 20 iterations in a 4-group setting with one background variant and 3 foreground variants. The x-axis represents each of the variants, and the y-axis represents the number of steps the pole was kept upright. We find that FQI performs slightly better for $c=1, 5$, but that it falters for the extreme variants. On the other hand, CFQI achieves comparable performance for all four variants. Error bars represent standard error. }
  \label{fig:mgnfqi}
\end{figure}

Now, we investigate the use of our method in a multi-group Cartpole setting. Recall in the two group setting that the background group was the original Cartpole setup and the foreground group had an added force pushing the cart to the left. Shifting to a multi-group paradigm, we now use three foreground groups in addition to the background group, where each of the groups has a distinct magnitude of force pushing the cart to the left. CFQI is appropriate for a multi-group setting because it avoids the retraining of $g_b$; in a multi-group setting, we only need to learn $g_b$ once. We pick $c=1, 5, 8$ for the foreground groups, and maintain that the background group is the original Cartpole. Our network structure is then modified to contain one set of background parameters and three sets of foreground parameters. To perform inference using CFQI, we pass a sample from a given group through both the shared parameters and a group-specific head if it is a foreground sample. For FQI, we adjust the architecture to use the same number of parameters as CFQI, but only use one set of parameters. Our results indicate that across CFQI can more successfully keep the pole upright than FQI across all variants (Figure \ref{fig:mgnfqi}). 

\newpage
\section{MIMIC-IV dataset}
\label{appendix:mimic}
The MIMIC-IV EHR dataset gathers records between $2008$ and $2019$ from the Beth Israel Deaconess Medical Center in Boston, Massachusetts. MIMIC-IV consists of de-identified data from over $524,000$ distinct hospital admissions from over $257,000$ unique patients~\citep{mimicivdataset}.

We filter the patient cohort by first selecting patients who were admitted to the ICU, were at least 18 years of age, and received at least one instance of potassium repletion. From this set, we select a foreground cohort of renal patients with end-stage renal disease (ESRD) ($n = 2,258$) using the corresponding ICD-9/ICD-10 codes, and a background cohort of 
non-renal patients, who received potassium but have functioning kidneys $(n = 36,023)$. To create patient trajectories, we discretize each patient's data into six-hour intervals, following previous RL approaches for this problem~\citep{Prasad_2020,copecat}. Each six-hour interval is considered one sample; each sample contains data from 56 covariates that describe the patient's vitals, labs, and prescribed medicines during that interval, as well as static metadata, such as comorbidities, age, and gender 
(Table \ref{erep_features}). We follow the imputation strategies proposed in earlier work~\citep{copecat}.
After discretizing patient trajectories into 6-hour intervals, the renal patients have on average 23 time-steps and the non-renal patients have on average 20 time-steps, indicating that renal patients spent on average more time in the ICU (18 hours more). 
Table \ref{erep_features} contains a list of all patient characteristics used to represent a state in our experiments.

Performing an initial exploration of our selected cohort, we find that renal patients have higher levels of creatinine (on average 280\% greater than the non-renal patients' creatinine level), which typically indicates abnormal kidney function~\citep{creatinine}. Additional factors that can affect potassium repletion include the presence of Coronary Artery Disease (CAD) and Rhabdomyolysis~\citep{cad, rhabdo}; the foreground and background cohort have comparable rates of both (16\% and 11\% respectively for CAD and 0.04\% and 0.09\% for Rhabdomyolysis).

The Markov decision process (MDP) for the MIMIC-IV environment is a tuple $(\mathcal{S}, \mathcal{A}, P, R)$. Here, $\mathcal{S}$ is a 56-dimensional vector containing information corresponding to each feature in Table \ref{erep_features}; $\mathcal{A} = \{[0, 1, 2, 3, 4, 5, 6]\}$ is the action space, each action corresponding to no repletion, three levels of low repletion and three levels of high repletion; $P$ contains the (unknown) state transition probabilities guiding a patient's state dynamics during their hospital visit; and $R$ is a reward function, which returns a value based on whether the patient's observed potassium level is outside the normal potassium range, and the cost of repletion. $R$ is the sum of a vector of length three; each element in this vector corresponds to the cost of intravenous repletion, a penalty for the patient's potassium level being too high, and a penalty for the patient's potassium level being too low, respectively. 
The reward function can be written as $r_t = w \cdot \phi_t(s_t, a_t, s_{t + 1})$ where $\phi$ is a vector function of length three such that:
\begin{align*}
    \phi_t(\cdot) &= 
    \begin{bmatrix}
    -\mathds{1}_{{a_t}^{route}[intravenous]} \\
    -\mathds{1}_{s_{t+1}[K] > K_{max}} \cdot 10 \left( 1 + e^{-\sigma (K - K_{max} - 1)} \right)^{-1} \\
    -\mathds{1}_{s_{t+1}[K] < K_{min}} \cdot 10 \left[1 - \left( 1 + e^{-\sigma (K - K_{min} + 1)} \right)^{-1}\right] \\
    \end{bmatrix} \\
    &\in
    \begin{bmatrix}
    \{0, -1\} \\
    (-10, 0) \\
    (-10, 0) \\
    \end{bmatrix}.
\end{align*}
Here, $K$ is the known measurement of potassium, and $K_{max}$ and $K_{min}$ define the upper and lower bounds of the target potassium range, respectively. For our experiments, we use $w=[5, 1, 1]$. To train our algorithm, like in the Cartpole environment, we use a SGD optimizer. We train for 500 epochs, with a constant learning rate of $1e-3$. The convergence criteria is the same as the Cartpole environment: the first stage of training culminates after three correct predictions are made on the background dataset, and the second stage terminates after three correct predictions are made on the foreground dataset. 

We report results using an importance sampling based log estimate of the value of the policy. We have an effective sample size of $1016$ on the renal dataset and $7657$ on the non-renal dataset. We use importance sampling by first learning a behavior policy $\pi_b$, and calculating the importance weight of a given sample as 
\begin{equation}
    \rho_T = \prod_{T=0}^T \frac{\pi_e(a_t|s_t)}{\pi_b(a_t|s_t)}
\end{equation}
where $\pi_e$ is the policy we are evaluating, and T is the length of a given trajectory. Then, the value of $\pi_e$ is 
\begin{equation}
    \hat{V}_{IS}(\pi_e) = \frac{1}{N} \sum_{i=1}^N \rho_{T-1}^i \sum_{t=0}^{T-1} \gamma^t r_{t+1}
\end{equation}
where $N$ is the total number of trajectories, $\gamma$ is a discount factor, and $r_{t+1}$ is the reward for the next sample in the trajectory. This importance sampling estimate follows from earlier work~\citep{Prasad_2020}.
\begin{table*}[ht]{%
\small
\begin{tabular}{|c|c|}
\hline
     \textbf{Feature}        & \textbf{Description}\\
    \hline
Age &  Age of patient at admission \\\hline
Gender & Patient gender  \\\hline
Patient Weight  & Patient Weight (kg)  \\\hline
Length of Stay  & Length of patient's stay (days)  \\\hline
Heart Rate & Heart Rate (bpm) \\\hline
Respiratory Rate & Rate of breathing (breaths per minute)\\\hline
Oxygen Saturation & Oxygen saturation in blood (\%) \\\hline
Temperature & Body temperature (\degree F) \\\hline
Systolic BP & Systolic Blood Pressure (mmHg)\\\hline
Diastolic BP & Diastolic Blood Pressure (mmHg) \\\hline
Potassium (IV) & Potassium administered through IV (mL), and \# hours administered \\\hline
Potassium (non-IV) & Potassium administered orally (mg), and \# hours administered \\\hline
Potassium & Potassium measured in blood \\\hline
Calcium & Calcium measured in blood \\\hline
Alanine Transaminase & Alanine Transaminase measured in blood \\\hline
Calcium (IV) & Calcium administered through IV (mL) \\\hline
Calcium (non-IV) & Calcium administered orally (mL) \\\hline
Chloride & Chloride measured in blood \\\hline
Phosphate (IV) & Phosphate administered through IV (mL), and \# hours administered \\\hline
Phosphate (non-IV) & Phosphate administered orally (mg), and \# hours administered \\\hline
Phosphate & Phosphate measured in blood \\\hline
Magnesium (IV) & Magnesium administered through IV (mL), and \# hours administered \\\hline
Magnesium (non-IV) & Magnesium administered orally (mg), and \# hours administered \\\hline
Magnesium & Magnesium measured in blood \\\hline
Sodium & Sodium measured in blood \\\hline
Vasopressors & Administered to constrict blood vessels \\\hline
Beta Blockers & Administered to reduce blood pressure \\\hline
Loop Diuretics & Administered to treat hypertension, edema \\\hline
Insulin & Administered to promote absorption of glucose from blood \\\hline
Dextrose & Administered to increase blood sugar \\\hline
Oral Nutrition & Orally administered nutrition supplements \\\hline
Parenteral Nutrition & Non-orally administered nutrition supplements \\\hline
Dialysis & Binary indicator of dialysis procedure \\\hline
Coronary Artery Disease & Binary indication of disease \\\hline
Atrial Fibrillation & Binary indication of disease \\\hline
Congestive Heart Failure & Binary indication of disease \\\hline
End-stage Renal Disease & Binary indication of disease \\\hline
Rhabdomyolysis & Binary indication of disease \\\hline
Sarcoidosis & Binary indication of disease \\\hline
Sepsis & Binary indication of disease \\\hline
Anion Gap & Measured in blood \\\hline
Blood Urea Nitrogen & Measured in blood \\\hline
Creatine Phosphokinase & Measured in blood \\\hline
Hemoglobin & Measured in blood \\\hline
Glucose & Measured in blood \\\hline
Creatinine & Measured in blood \\\hline
Lactic Acid Dehydrogenase & Measured in blood \\\hline
White Blood Cell & Count measured in blood \\ \hline
Red Blood Cell & Count measured in blood \\ \hline
Urine & Total volume of urine expelled (mL)\\ \hline
Expired & Indicator of whether the patient died in the hospital \\ \hline
\end{tabular}
}
\caption{Features used to recommend electrolyte repletion.}
\label{erep_features}
\end{table*}

\end{document}